%% file: templateArxiv.tex
\documentclass{article}

\usepackage{PRIMEarxiv}

\usepackage[utf8]{inputenc} 
\usepackage[T1]{fontenc}    
\usepackage{hyperref}       
\usepackage{url}            
\usepackage{booktabs}       
\usepackage{amsfonts}       
\usepackage{nicefrac}       
\usepackage{microtype}      
\usepackage{lipsum}
\usepackage{fancyhdr}       
\usepackage{graphicx}       
\usepackage{xcolor}
\usepackage{multirow}
\usepackage[section]{placeins}

\graphicspath{{media/}}     

\title{An Analysis of Language Frequency and Error Correction for Esperanto
}

\author{
  Junhong Liang \\
  Institute of Automation, Chinese Academy of Sciences \\
  School of Artificial Intelligence, University of Chinese Academy of Sciences  \\
  \texttt{liangjunhong2022@ia.ac.cn} \\
}

\begin{document}
\maketitle

\begin{abstract}

Current Grammar Error Correction (GEC) initiatives tend to focus on major languages, with less attention given to low-resource languages like Esperanto. In this article, we begin to bridge this gap by first conducting a comprehensive frequency analysis using the Eo-GP dataset, created explicitly for this purpose. We then introduce the Eo-GEC dataset, derived from authentic user cases and annotated with fine-grained linguistic details for error identification. Leveraging GPT-3.5 and GPT-4, our experiments show that GPT-4 outperforms GPT-3.5 in both automated and human evaluations, highlighting its efficacy in addressing Esperanto's grammatical peculiarities and illustrating the potential of advanced language models to enhance GEC strategies for less commonly studied languages.
\end{abstract}

\keywords{Grammar Error Correction \and Esperanto}

\section{Introduction} \label{sec:intro}

Current Grammar Error Correction (GEC) systems predominantly target major languages like English\cite{bryant-etal-2017-automatic,fan2023grammargpt,bryant2023grammatical}, Chinese\cite{zhang-etal-2022-mucgec,zhang-etal-2023-nasgec}, German\cite{boyd-2018-using} and Japanese\cite{Liu2018AutomaticEC}. This focus is driven by the availability of comprehensive datasets and the specific linguistic characteristics inherent to these languages. Consequently, the exploration of GEC methodologies for low-resource languages has been largely overlooked, leaving a significant gap in the analysis and development of error correction strategies for these less-studied languages.

Recently, Large Language Models (LLMs) have revolutionized the field of Natural Language Processing (NLP) by equipping these models with the ability to generate text that close to human language. LLMs have attracted considerable attention for their proficiency in English language tasks. Recent studies, however, reveal their potential across various languages. Despite this broad applicability, our analysis identifies a notable gap in the research landscape, particularly concerning Esperanto. As a constructed language, Esperanto presents unique challenges in terms of frequency distribution and grammar error correction that have yet to be thoroughly explored.

This article delves into the word and letter frequency specific to Esperanto and embarks on a preliminary investigation into the capabilities of GPT-3.5 and GPT-4—innovations by OpenAI\footnote{\url{chat.openai.com}} —in correcting Esperanto grammar errors. Our contributions are as follows:
\begin{itemize}
    \item We propose the Eo-GP dataset which is selected from the Gutenberg Project to analyze the word and letter frequency of Esperanto.
    \item We collected the Eo-GEC dataset which contains the authentic grammar errors made by language learners.
    \item We quantitative and qualitatively evaluated LLMs as a GEC tool for the Esperanto language.

\end{itemize}

\section{Related Works}
\subsection{Esperanto Resources}
Low-resource languages, such as Esperanto, present unique challenges in multilingual NLP research due to the limited availability of comprehensive datasets. While numerous multilingual datasets exist, the inclusion of Esperanto remains surprisingly sparse. Existing Esperanto datasets primarily cater to language learners, lacking the scope and depth required for comprehensive research purposes. 
\paragraph{Educational Resources}
For language learners, there are several platforms available. Lernu \footnote{\url{https://lernu.net/esperanto}} offers a comprehensive language course, a dictionary, and a forum dedicated to Esperanto. Edukado.net \footnote{\url{https://edukado.net/}} serves as an integrated platform catering to Esperanto teachers and learners, providing access to important Esperanto events. Project Gutenberg \footnote{\url{https://www.gutenberg.org/ebooks/bookshelf/34}} offers a wide range of Esperanto books for readers. Additionally, Esperantistoj \footnote{\url{https://esperanto.lodestone.org/ligiloj/en}} provides a collection of useful links related to media, language learning, and travel.
\paragraph{NLP Research Resources}
Most of the resources used for NLP research in Esperanto are mainly a part of multilingual datasets. Goyal  builds Commoncrawl Corpus in 100 languages and then uses a language identification model to select Esperanto contexts\cite{wenzek-etal-2020-ccnet}, which is later combined with Wikipedia Esperanto dataset to train multilingual model\cite{conneau-etal-2020-unsupervised}. Lepzig Corpora Collection \cite{goldhahn-etal-2012-building} introduces an automatic collection method for low-resource languages and includes online source data for Esperanto. OSCAR \cite{abadji-etal-2022-towards} provides a collection of unannotated raw data and devises a method to improve the quality of raw data from Common Crawl \footnote{\url{https://commoncrawl.org/}}.

\subsection{Esperanto-Centric Research}
There are few researches which focus on the inherent characteristic of Esperanto language itself. 
Early works focus on the translation of Esperanto texts. Ausloos investigates the effect of punctuation on sentence length and writing style, proposing that sentences seem to be more reliable than word distributions in discussing an author style.\cite{DBLP:journals/corr/abs-1004-4848}. A different split method enable the utilization of several byte-pair encoding models and could improve the performance of English-Esperanto translation task. \cite{DBLP:journals/corr/abs-2011-14190}, a bilingual dictionary between two language we built by aligning monolingual word embedding spaces in an unsupervised way\cite{DBLP:journals/corr/abs-1710-04087}.

\subsection{Grammar Error Correction}

\paragraph{Deep Learning Models} Grammatical Error Correction (GEC) is the task of automatically detecting and correcting errors in text\cite{bryant2023grammatical}. There are mainly two approaches towards GEC, the first approach is to treat GEC as a monolingual translation task\cite{yuan-briscoe-2016-grammatical} where sequence-to-sequence models such as BART\cite{lewis2019bart} are used. Another approach treats GEC as sequence tagging task, for each input token, a sequence tagging (such as "KEEP", "DELETE", "ADD") is generated, then a post-processing method is used to generate the correct sentence. This approach is used in models such as BiLSTM-CRF\cite{liu2018detecting} and GECToR\cite{omelianchuk2020gector}.

\paragraph{Annotation and Evaluation} Dahlmeier proposes M2Scorer to evaluate the quality of GEC by calculating the phrase-level alignment between the source and target sentence\cite{dahlmeier-ng-2012-better}. Bryant puts forward a fine-grained linguistic annotation scheme as well as an evaluation criterion, ERRANT, and automatically extract the edits that transform the source to the target sentence\cite{bryant-etal-2017-automatic}. Based on the previous works on English GEC, Zhang adapts ERRANT for Chinese GEC and proposes a multiple-reference multiple-resource Chinese GEC dataset and the evaluation method, ChERRANT\cite{zhang-etal-2022-mucgec}.  Uz provided an automatic GEC annotation method for Turkish \cite{uz-eryigit-2023-towards}.

\paragraph{The Application of LLMs} Recent advancements in Large Language Models (LLMs) have significantly improved the ability of GEC systems\cite{bryant2023grammatical}.  GrammarGPT\cite{fan2023grammargpt} employs a hybrid human-annotated and GPT-generated dataset for Chinese GEC tasks. Adding explanation after error correction could help users to have a deeper understanding, Grammar Error Explanation (GEE) is proposed to bridge this gap \cite{song2023gee}. ChatGPT has excellent error correction abilities and tends to play freely and generate a more fluent output\cite{fang2023chatgpt}. 

\section{Frequency Analysis}

\subsection{Letter Frequency Analysis using La Eta Princo}
\begin{figure}
  \centering
  \includegraphics[width=\linewidth]{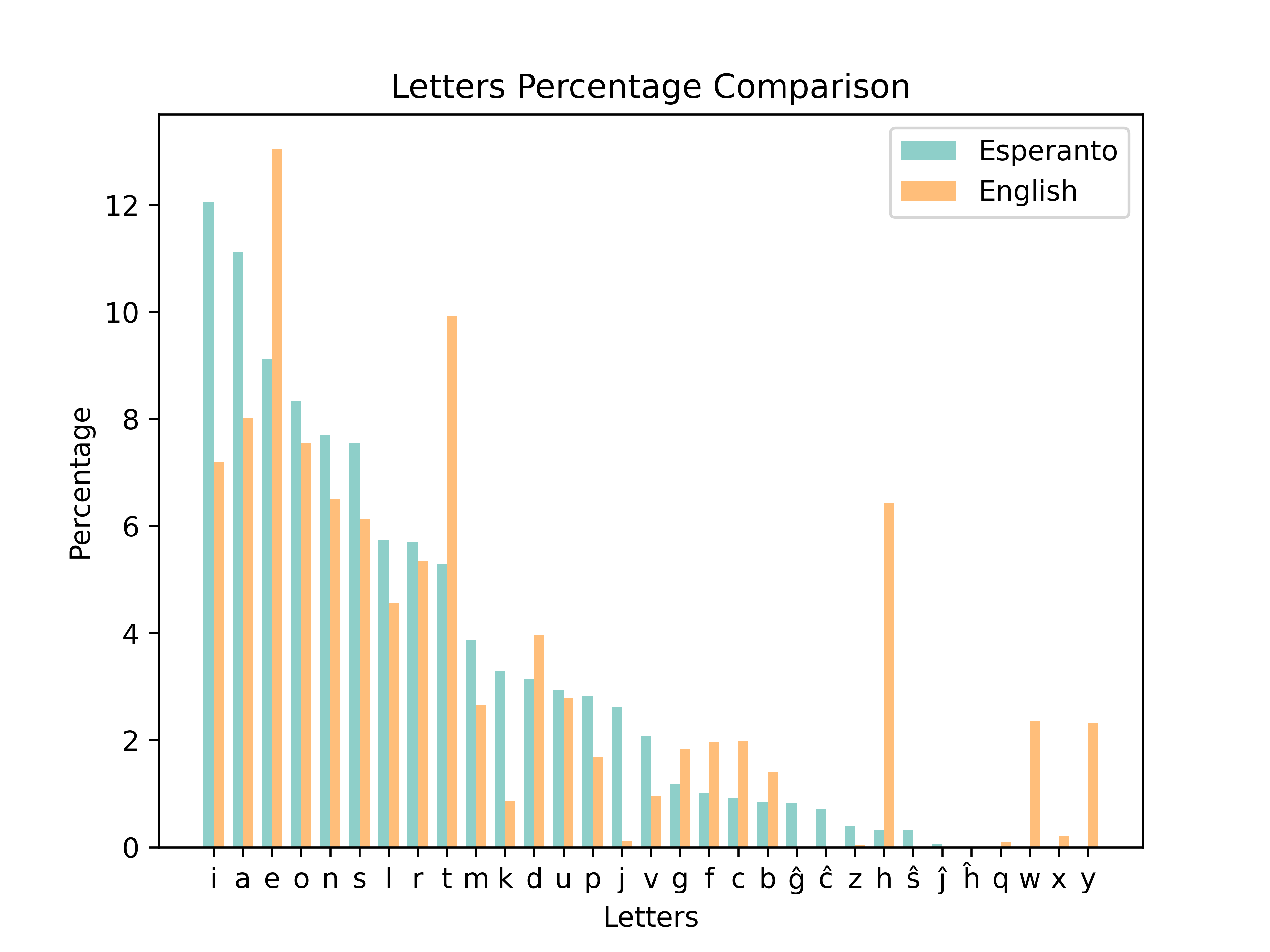}
  \caption{The comparison of alphabet frequencies between Esperanto and English version of La Eta Princo}
  \label{fig:esperanto_english_le_eta_princo}
\end{figure}

A complete set of Esperanto letters is demonstrated in Table \ref{tab:esperanto_alphabet} in Appendix \ref{sec:eo-alphabet}. We select Le Petit Prince by Saint Antoine-Exupery and analyze the alphabet frequency of Esperanto version (La Eta Princo) and English version (Le Petit Prince). The result is shown in  Figure \ref{fig:esperanto_english_le_eta_princo}. 

In the comparative analysis of letter frequency between the Esperanto and English translations of "Le Petit Prince", we observe several distinct discrepancies that merit attention. These variations can be ascribed to a multitude of linguistic factors intrinsic to each language's structural and orthographic conventions.

\paragraph{Phonetic composition of Esperanto} Esperanto is engineered for phonemic orthography, ensuring that each grapheme corresponds unequivocally to a single phoneme. This one-to-one correspondence aims for a uniform distribution of letter usage. Conversely, English orthography is characterized by its depth, with a multitude of irregular phoneme-to-grapheme correspondences, which results in a less predictable letter frequency distribution. Esperanto orthography is meticulously phonetic, where each letter corresponds to a single sound, and each sound is determined by a fixed combination of letters. Consequently, the letter 'k' in Esperanto consistently represents the voiceless velar plosive sound [k], a sound which in English can be denoted by both 'c' (as in "cat") and 'k' (as in "kite"). The usage of 'k' in Esperanto is in alignment with its objective for linguistic simplicity and regularity, thus eliminating the ambiguity found in English spelling conventions. This results in a higher frequency of "k" in Esperanto than in English.

\paragraph{Lexical Foundation} Esperanto's lexical foundation, amalgamating elements from Romance, Germanic, and Slavic lexicons, is regularized both grammatically and lexically, thereby affecting letter frequency. English, a language of Germanic origin with substantive Romance language accretion, particularly from French and Latin, exhibits a different pattern of letter usage reflective of its etymological diversity. For example, in Esperanto language, the "question pronouns” start with "ki", as well as for relative pronouns, this result in a higher percentage of "k" in Esperanto than in English. 

\paragraph{Agglutinative morphology} Esperanto's morphology is predominantly agglutinative, relying on consistent affixation to construct meaning, which may result in an elevated occurrence of certain letters. English employs a more heterogeneous morphological system, blending inflectional and derivational processes, which influences letter frequency in a disparate manner. We notice that "j" has a much higher percentage in Esperanto rather in English, we believe that there might be several reasons. 1) Forming the plural of nouns is generally straightforward by adding "j" to the end of a noun. 2) Meanwhile, adjectives should agree with nouns by adding "j" to the end of an adjective. 3) Some affixes of Esperanto include letter "j", for example, "-ejo" (a place characterized by the root word, such as malsanulejo, hospital) and "-ujo" (a container, a country, such as monujo (purse))

\subsection{EO-GP Dataset} We select Esperanto books in Gutenberg Project\footnote{\url{https://www.gutenberg.org/ebooks/bookshelf/34}}, which provides a readable HTML e-book version for a part of Esperanto books. We then download all the documents on the webpage and extract HTML files for processing. The process is demonstrated as follows:
\begin{itemize}
    \item Choose Esperanto Documents. For each HTML file, an indicator of language is presented at the beginning. Noticing that some books use English or French to teach Esperanto or create bilingual dictionaries, we only select Esperanto books, which means that the language should be completely Esperanto. 
    \item Process HTML File. We use \verb|BeautifulSoup|\footnote{\url{https://pypi.org/project/beautifulsoup4/}} to process HTML files and extract Esperanto paragraph texts (marked by \verb|<p>| and \verb|<\p>|) 
    \item Save the documents. Finally, we save the documents in \verb|txt| format.
\end{itemize}
The original Gutenberg Project page includes 107 Esperanto book archives, finally we filter out 8 which are not written in Esperanto language. Finally, we create our dataset based on 99 books. 

\paragraph{Entropy of Esperanto Letters}
We calculate the entropy of Esperanto letters using Eo-GP dataset by the following formula

\begin{equation}
    H(X) = -\sum_{i=1}^{|X|} P(x_i) \log_2 P(x_i)
\end{equation}

where $|X|$ indicates the number of lowercase Esperanto letters. We use the statistics in "La Eta Princo" and the entropy of Esperanto letters is 4.14, while the entropy for English letters is 4.18. This indicates that Esperanto letters may exhibit slightly less unpredictability or variability in their occurrence compared to English letters. This can be interpreted as Esperanto having a slightly more regular or predictable letter distribution.

\subsection{Word Frequency Analysis}
We employ EO-GP dataset which is collected from 99 books in Gutenberg Project. We firstly read the \verb|txt| files, next, we split the spaces to get the words. Due to the relatively stable word endings in Esperanto, we utilize the rules governing word endings to determine whether a given word is a part of the Esperanto vocabulary. Finally, we count the word frequency based on the word lists. 

\begin{figure}[h]
  \centering
  \includegraphics[width=0.9\linewidth]{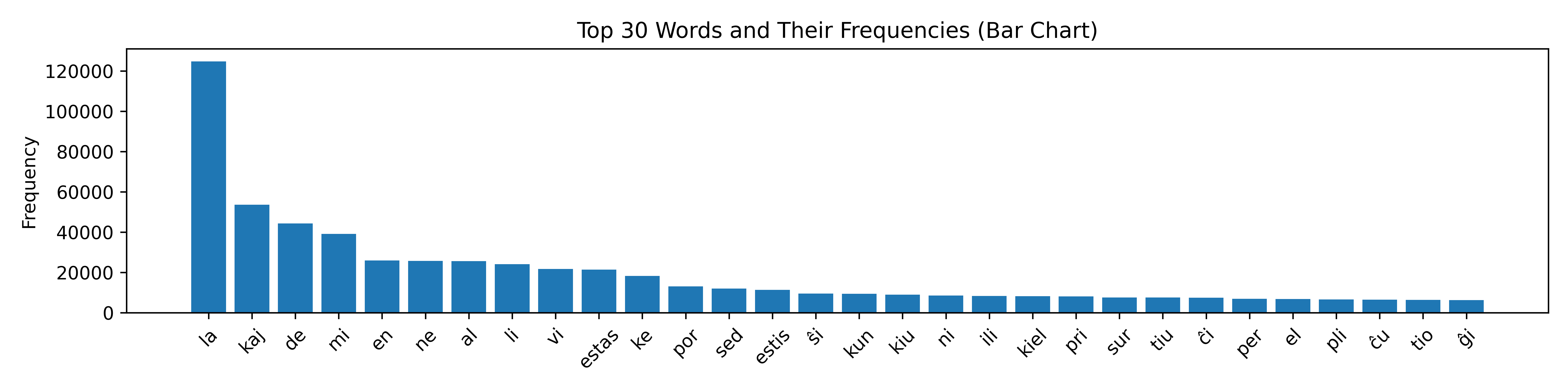}
  \caption{Top 30 Esperanto word frequencies in EO-GP}
  \label{fig:esperanto_word_freq}
\end{figure}

The result of top 30 word frequency is shown in Figure \ref{fig:esperanto_word_freq}. It can be observed from the figure that the frequency of word occurrences aligns with Zipf's law, the frequency of a word is inversely proportional to its rank or position in a frequency distribution. A small set of words, often referred to as "stop words," such as articles and prepositions, dominate language usage, while a vast number of less common words make up the tail end of the distribution. The words with higher frequency primarily consist of articles, pronouns, conjunctions, and prepositions such as ``la" (the), ``kaj" (and), ``de" (of), ``mi" (I) and ``en" (at) 

\begin{figure}[h]
  \centering
  \includegraphics[width=0.9\linewidth]{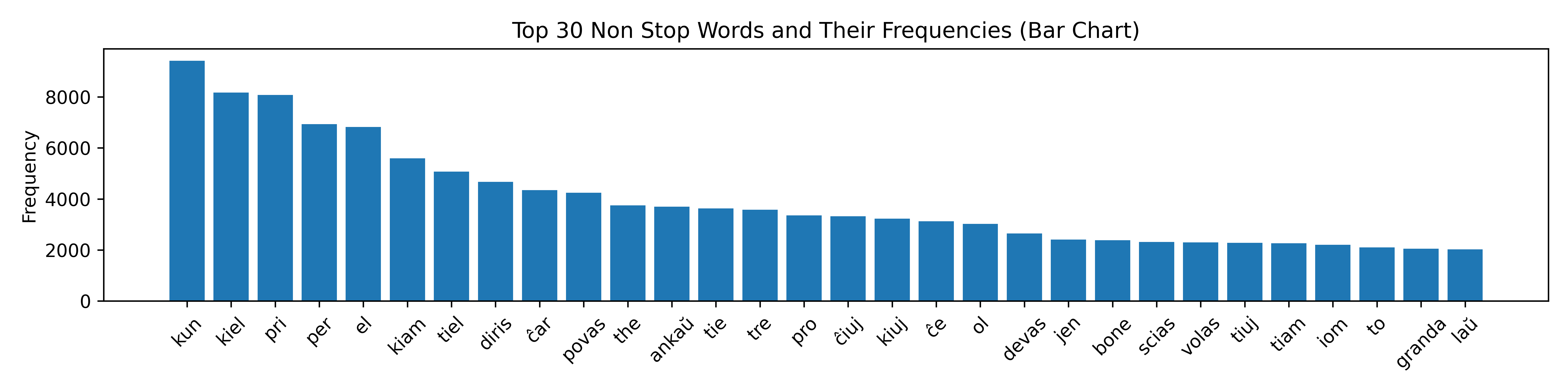}
  \caption{Top 30 Esperanto non-stop word frequencies in EO-GP}
  \label{fig:esperanto_word_freq_non_stop}
\end{figure}

We then use the Esperanto stop words dictionary provided by stopwords-iso \footnote{\url{https://github.com/stopwords-iso}} and we filter out the stop words in the counting. We plot the frequency of top 30 non-stop words in Figure \ref{fig:esperanto_word_freq_non_stop}.

\subsubsection{Fitting Zipf's formula}
We selected top 100 words in all collected words as well as in non-stop words and plot their log frequency along with log rank.
\begin{figure}[h]
  \centering
  \includegraphics[width=0.9\linewidth]{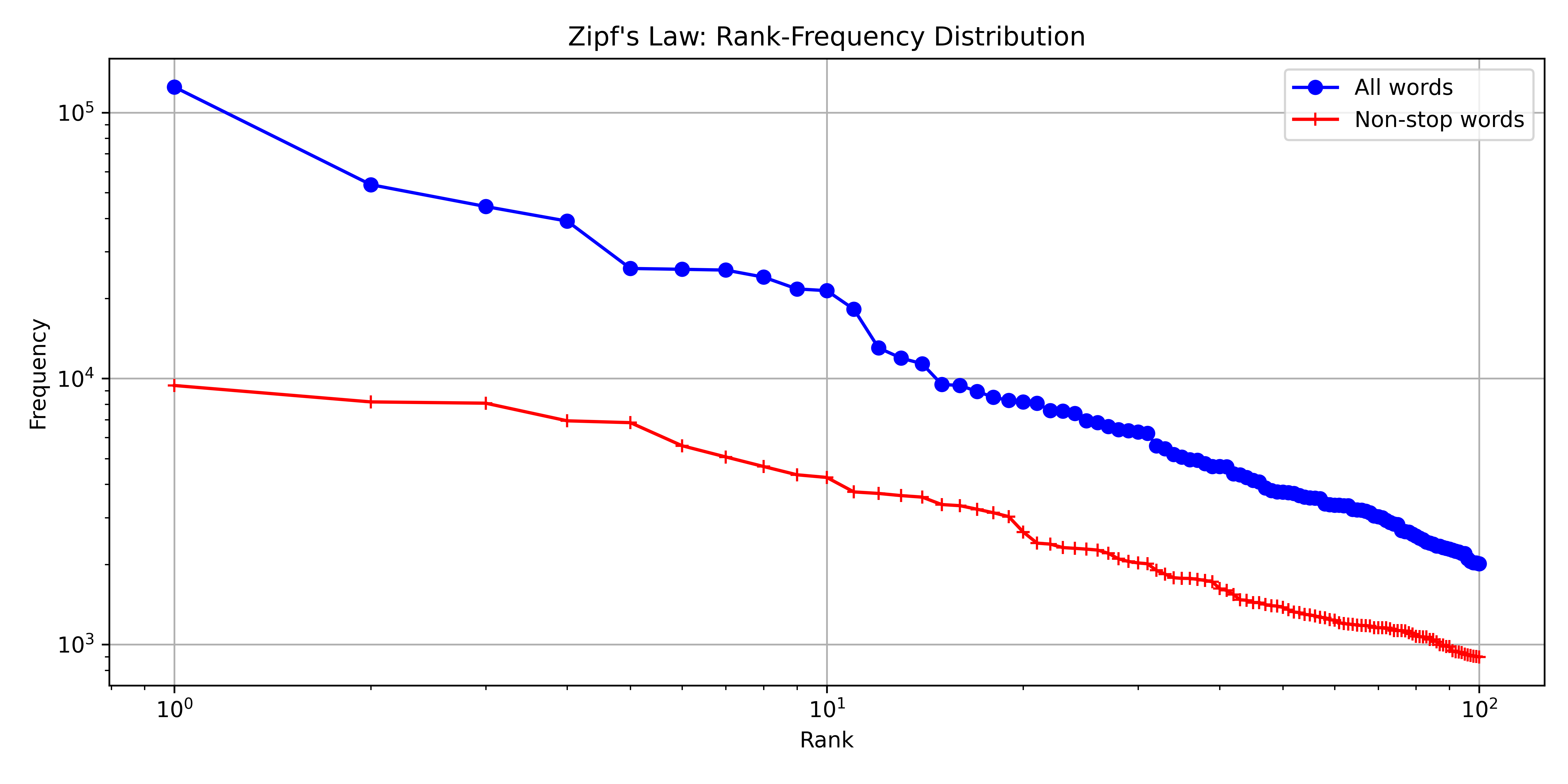}
  \caption{The logarithmic frequency-log rank plot of the top 100 words and non-stop words.}
  \label{fig:Zipf's law}
\end{figure}

We fit the plots using Zipf-Mandelbrot law:
\begin{equation}
\label{eqn:zipf-mandelbrot}
    f(w) = \frac{k}{(r(w)+b)^a}
\end{equation}
where $f(w)$ and $r(w)$ represents the frequency and rank of word $w$ respectively, $k$,$a$,$b$ are fitted parameters.  If only fitting the inverse proportionality law between frequency and rank, we can assume that $b=0$. Logarithms are applied to both sides of the equation concurrently, then equation \ref{eqn:zipf-mandelbrot} becomes
\begin{equation}
\label{eqn:zipf-mandelbrot-log}
    \log f(w) = \log k - a \log (r(w))
\end{equation}
Therefore, the expressions $\log f(w)$ and $\log r(w)$ can be represented by a straight line with a slope of $-a$.

We fit the data points in Figure \ref{fig:Zipf's law} using equation \ref{eqn:zipf-mandelbrot-log}. The result is shown in Table \ref{tab:zipf_fitting}. After performing the fitting process, it is evident that a linear pattern is significantly established. The data points exhibit a strong alignment along a straight line, indicating a high degree of correlation between the log-frequency and log-rank. This finding verifies Zipf's law. The goodness-of-fit measures, such as the coefficient of determination (R-squared), further confirm the robustness of the linear fit.
\input{tables/zipf_fitting}

\section{Error Correction}
\subsection{Esperanto error types}\label{sec:error_type}
In Appendix \ref{apdx:list_of_error_type}, a comprehensive compilation of Esperanto error type codes is provided. It is important to note that each error code can be represented using two or three components, with each component separated by a colon. The initial component is designated by the letters "M," "U," or "R," signifying missing, unnecessary, or redundant error types, respectively. This classification enables the evaluation of error correction capabilities associated with each type. The second component denotes the Part of Speech (PoS) of the error token, in the case of PoS errors and morphology errors, or specifies the error type for other error categories. It is worth mentioning that only morphology error codes encompass a third component, which articulates the particular error sub type related to the PoS. Hence, every code can be represented using either two or three components.

Then, in Table \ref{tab:example_of_errors}, we list the existing major error types in our Eo-GEC dataset.

Different from English, Esperanto features a distinctive group of words, commonly referred to as "table words", that can be systematically arranged based on their prefixes and suffixes. These words, essential for forming interrogatives and clauses, are constructed using one of five prefixes (\textbf{i-}, \textbf{ki-}, \textbf{ti-}, \textbf{ĉi-}, and \textbf{neni-}) and one of nine suffixes (\textbf{-a}, \textbf{-al}, \textbf{-am}, \textbf{-e}, \textbf{-el}, \textbf{-es}, \textbf{-o}, \textbf{-om}, \textbf{-u}). The precise meaning of each table word is determined by its specific combination of prefixes and suffixes, allowing for a high degree of predictability in their usage.

Moreover, table words are subject to inflectional changes to convey grammatical nuances. For instance, words ending in -a, -e, -o, and -u can incorporate the "n" suffix to signal the accusative case, indicating the object of an action. Similarly, the "j" suffix is used to denote plurality for words ending in -a and -u. This methodical structure underpins the predictability of table words' meanings. However, despite their systematic design, mastering the application of table words poses a considerable challenge for learners, reflecting the complexity of Esperanto's grammatical framework. In the annotation process, the table words errors have a special annotation marker "TABLE".
\subsection{Fine-grained Linguistic Annotation}
To evaluate the source sentences, we employ a fine-grained annotation scheme, manually identifying each erroneous token within a sentence. Following the identification process, we rectify the sentence by applying the appropriate error type and word transformation techniques outlined in Section \ref{sec:error_type}. To ensure consistency and accuracy in our annotations, we have established a set of specific rules governing the annotation process.
\begin{itemize}
    \item The annotation has two parts, a number indicating the error index of the original source sentence and an error type including the error type and the part of speech. 
    \item We first annotate the "Replace" errors, then "Unnecessary" and "Missing" errors. The index for unnecessary and missing errors is based on the source sentence. 
    \item If two words are changing the order, the annotation has three parts, index for the two words and error type "R:WO", if there is an orthography error that two words should be put together to form a word, this will also be annotated with three parts, and the error type is "R:ORTH".
    \item If the target sentence significantly diverges from the original sentence, the error type is "R:OTHER".
\end{itemize}
\begin{figure}[h]
  \centering
  \includegraphics[width=0.6\linewidth]{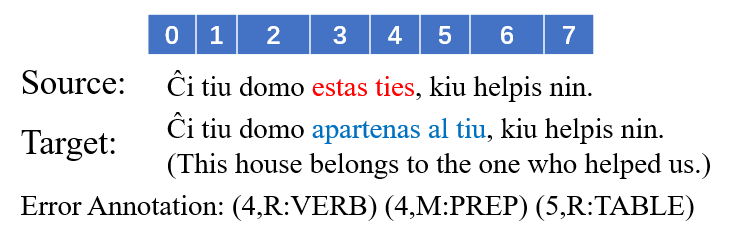}
  \caption{An Annotation Scheme for Esperanto Grammar Correction. The source and target sentence as well as the English translation are listed.}
  \label{fig:annotate}
\end{figure}

We list some major error type and give their descriptions and examples shown in Table \ref{tab:example_of_errors}
\input{tables/example_of_errors}

\subsection{A descriptive analysis of EO-GEC dataset}
Our dataset includes 307 grammatically wrong sentences and their corrections, as well as their fine-grained level linguistic annotations.

\begin{figure}[h]
  \centering
  \includegraphics[width=0.9\linewidth]{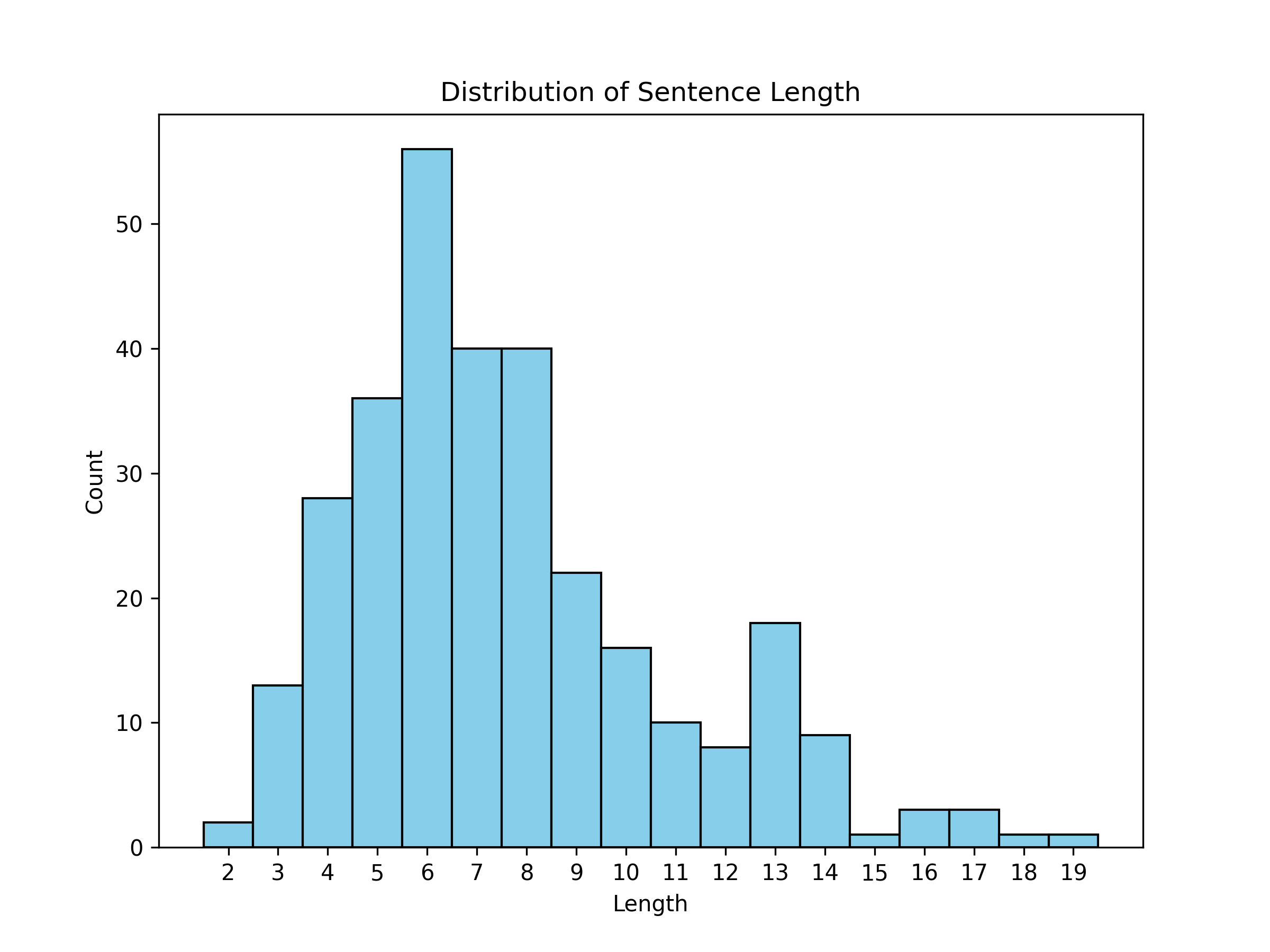}
  \caption{Length distribution of Esperanto sentences}
  \label{fig:length_stat}
\end{figure}

\paragraph{Length Statistics}
The word count for each sentence was summarized, and a histogram of the source sentences' lengths is presented in Figure \ref{fig:length_stat}. The average length of source sentences is 7.64 words, and for target sentences, it is 7.76 words. Given that the corpus predominantly comprises errors identified in users' Esperanto grammar exercises and grammar books, the sentences tend to be relatively short.
\paragraph{Source Analysis} The source for our Eo-GEC dataset is listed as follows.
\begin{itemize}
    \item \textbf{Teach Yourself Complete Esperanto} Teach Yourself Complete Esperanto is a textbook for Esperanto learners. It provides beginner to intermediate (B2 according to CEFR level) courses. We ask three Esperanto learners to summarize the mistakes they make in grammar practice questions. 
    \item \textbf{Plena Manlibro de Esperanta Gramatiko} Plena Manlibro de Esperanta Gramatiko\footnote{\url{https://bertilow.com/pmeg/elshutebla/index.html}} (PMEG, English: Complete Manual of Esperanto Grammar) is a book which explains Esperanto grammar in an easy-to-learn format. It was mostly written by Bertilo Wennergren and is for ordinary Esperanto speakers who want to study Esperanto's grammar, word construction, writing, and pronunciation. We collect the grammar mistakes labeled with an asterisk (*) in this book. 
    \item \textbf{User Writing} We collect the chatting record of the Esperanto community in the language learning platform HelloTalk, which provides an error correction system, allowing native speakers to correct the mistakes of Esperanto Learners. 
    \item \textbf{Nivelo al Nivelo} Nivelo al Nivelo (Level to Level) is an Espearnto standard test guide for the UEA-KER exam \footnote{\url{https://edukado.net/ekzamenoj/ker}}, we select the grammar questions from B1 and B2 level. 
\end{itemize}

\begin{figure}[!htbp]
  \centering
  \includegraphics[width=0.9\linewidth]{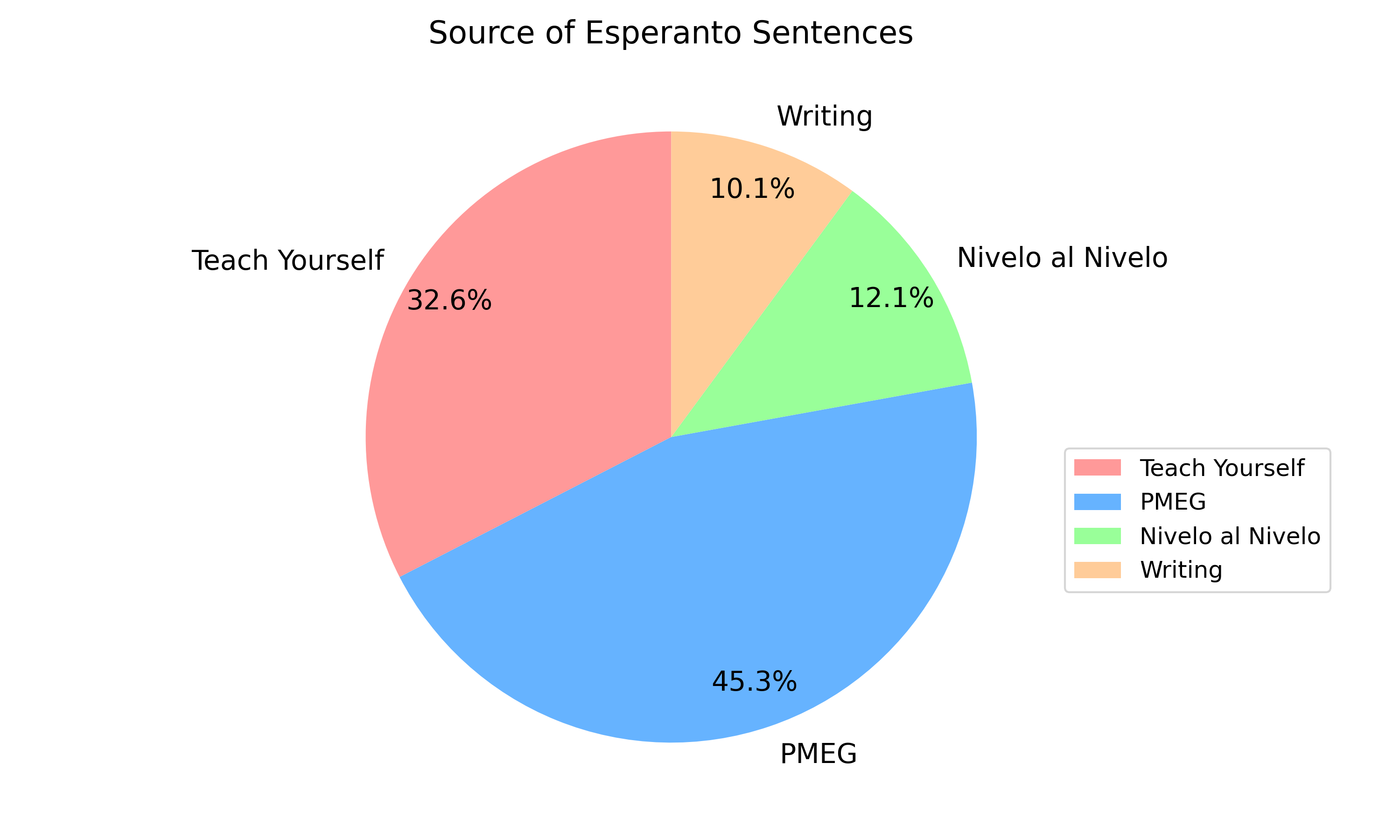}
  \caption{Source distribution of Esperanto sentences}
  \label{fig:source_stat}
\end{figure}

We analyze the source of the sentences shown in Figure \ref{fig:source_stat}. The selection of erroneous sentences in this study primarily relies on PMEG, as it offers an extensive list and clear grammatical rules regarding incorrect sentences. Meanwhile, the scarcity of information from written sources is attributed to two main factors: firstly, compared to other languages, Esperanto is a relatively constructed language with a smaller speaker base; secondly, communication in Esperanto often overlooks grammatical mistakes since the primary objective is to convey information to others. As a result, error correction is not very common on language exchange platforms such as HelloTalk.

\begin{figure}[!htbp]
  \centering
  \includegraphics[width=0.6\linewidth]{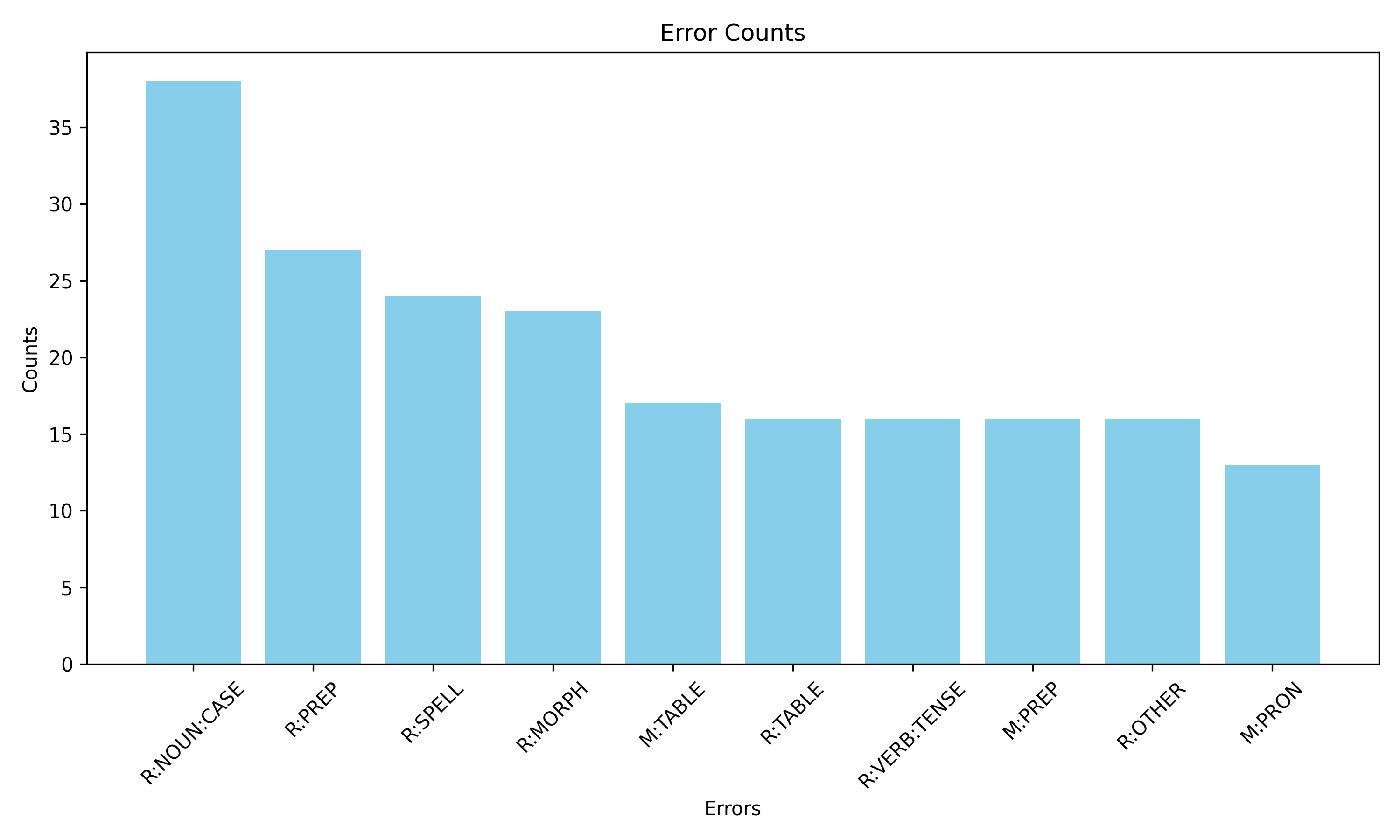}
  \caption{Top 10 error of Esperanto sentences}
  \label{fig:error_stat}
\end{figure}

\begin{figure}[!htbp]
  \centering
  \includegraphics[width=0.6\linewidth]{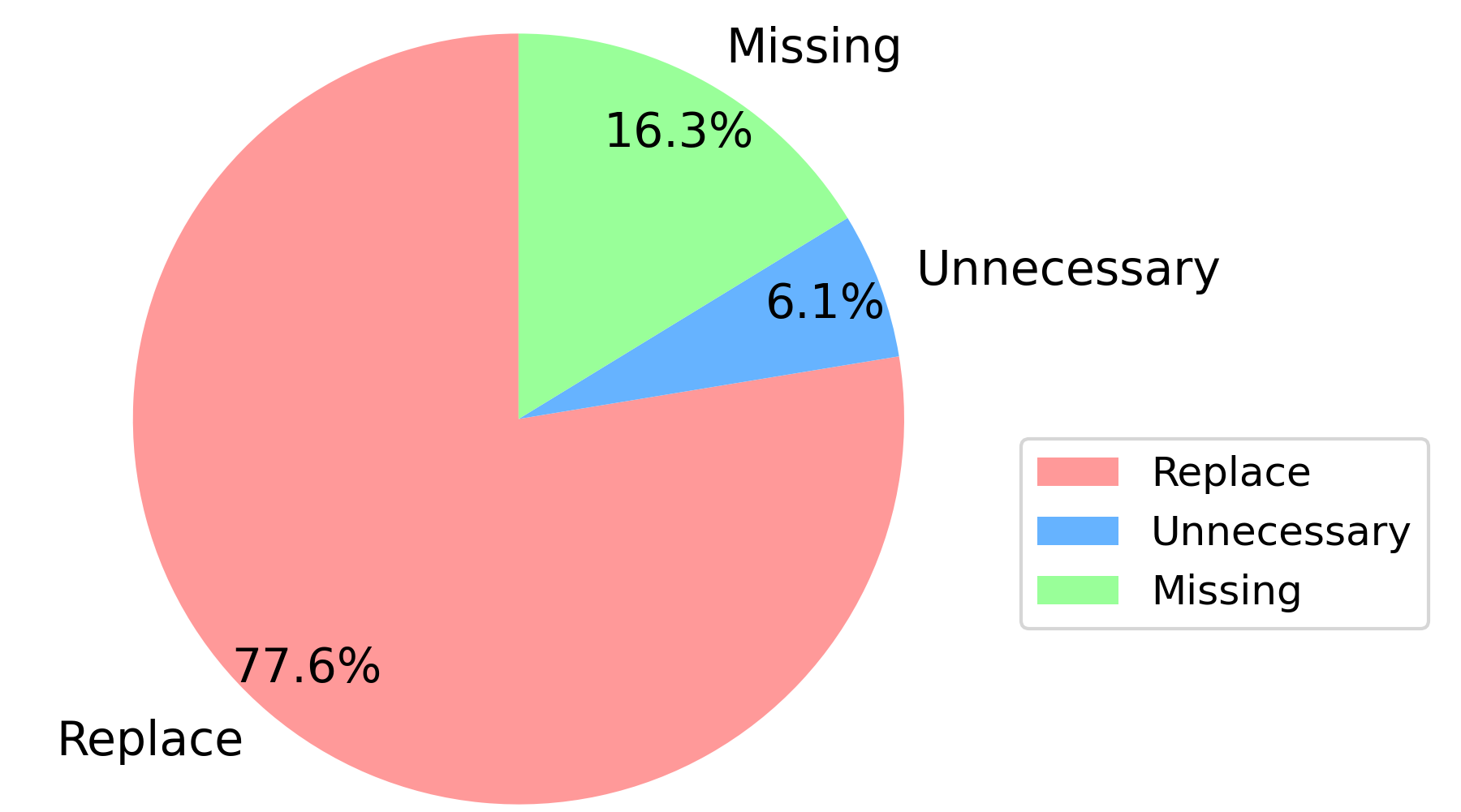}
  \caption{Pie Chart for the percentage of replace, missing and unnecessary errors.}
  \label{fig:rmu_stat}
\end{figure}

\begin{figure}[!htbp]
  \centering
  \includegraphics[width=0.6\linewidth]{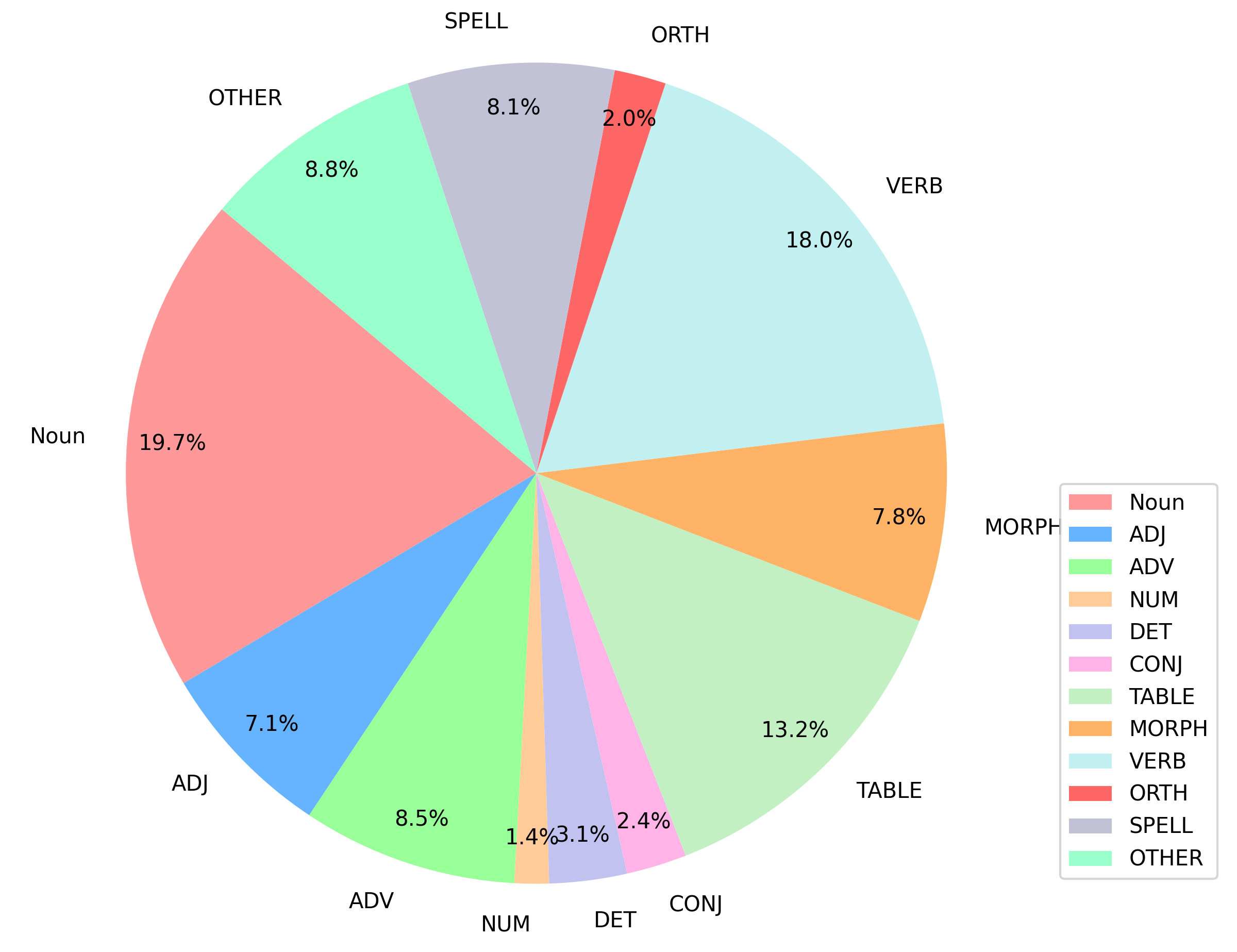}
  \caption{Pie Chart for part of speech of fine-grained linguistic errors}
  \label{fig:pos_stat}
\end{figure}

\paragraph{Error Type Ranking} We have also incorporated a detailed linguistic annotation of Esperanto sentences, categorizing and ranking their occurrences as depicted in Figure \ref{fig:error_stat}. The predominant error type identified pertains to nouns. Esperanto nouns exhibit two cases: nominative and accusative, a distinction that frequently leads to errors in grammar exercises and learner writings. Following closely, the misuse of prepositions ranks second. This confusion often stems from the similarity in spelling but the difference in meaning among some prepositions, such as pri (about), por (for), pro (because of), and po (per). Misunderstanding these prepositions is a common challenge in Esperanto KER (Komuna Ekzameno por la Rekono) tests. Thirdly, spelling errors are also prevalent, partly because Esperanto and English share many cognates that differ in spelling, leading learners to confuse them. For instance, it is common for learners to misspell "koktelo" (cocktail) as "kocktelo" or "coktelo", owing to the similar pronunciations of "c", "k", and "ck" in English.
\paragraph{Replace, Missing and Unnecessary Errors}
Additionally, we examine the overall distribution of replacement, missing, and unnecessary errors (denoted as R, M, U respectively) in Figure \ref{fig:rmu_stat}. Replacement errors constitute over three-fourths of the total observed errors. This predominance is attributed to the nature of grammar exercises and textbooks, which typically focus on correcting the inflection of words or phrases without altering the length of the text. Our analysis also reveals that unnecessary errors occur less frequently than missing errors.
\paragraph{Part of Speech Errors}
Our analysis extends to the categorization of errors by part of speech, as illustrated in Figure \ref{fig:pos_stat}. Notably, the most frequent errors involve nouns and verbs, reflecting their prevalence in Esperanto vocabulary. In addition, errors associated with table words (correlative words) are significantly more common than other types. This trend underscores the complexity of mastering table words, a challenging grammar aspect for learners.
\section{Large Langauge Models for Esperanto Error Correction}
\begin{figure}[!htbp]
  \centering
  \includegraphics[width=\linewidth]{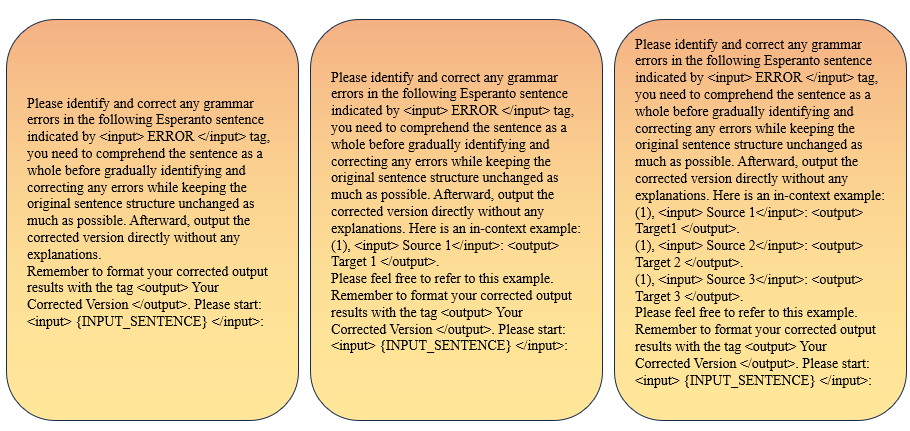}
  \caption{Prompts utilized for LLMs to correct Esperanto grammar errors across various settings: 0-shot (Left), 1-shot (Middle), and 3-shot (Right). Examples will replace 'Source' and 'Target' sentences, whereas 'INPUT\_SENTENCE' will be substituted with sentences containing grammatical errors in our dataset.}
  \label{fig:prompts}
\end{figure}

In this section, we employ both GPT-3.5 and GPT-4 to address and correct Esperanto grammar errors, utilizing specific prompts as illustrated in Figure \ref{fig:prompts}.

\subsection{Automatic Evaluation}

We utilize ERRANT\footnote{\url{https://github.com/chrisjbryant/errant?tab=readme-ov-file}} and M2Scorer\footnote{\url{https://github.com/nusnlp/m2scorer}} to evaluate the correction quality of the source sentences. The results are presented in Table \ref{tab:correction_res}. It is observed that both GPT-3.5 and GPT-4 achieve relatively satisfying results; Although GPT-4 does not score high on ERRANT compared to GPT-3.5, it outperforms in terms of M2Scorer.  

Notably, as the number of prompts increases, for GPT-3.5, the $F_{0.5}$ scores of ERRANT improve significantly. A substantial improvement in ERRANT and M2Scorer $F_{0.5}$ scores is evident between the 0-shot and 1-shot settings. However, for GPT-4, after 1-shot, scores slightly decrease with the increase in shots.

The best performance in M2Scorer evaluation is recorded at 1-shot using GPT-4. Interestingly, we discover there exists a mismatch in terms of $F_{0.5}$ scores for ERRANT and M2Scorer.

\input{tables/correction_quality}

\subsection{Human Evaluation}
Furthermore, we undertake a human evaluation of the corrected sentences by engaging three Esperanto evaluators, each possessing at least a KER B2 diploma. Two evaluators independently assess the quality of each revised sentence using three predefined quality markers:
\begin{itemize}
    \item \textbf{GOOD} The sentence has been accurately revised without significant alteration to its original meaning. 
    \item \textbf{ACCEPTABLE} The sentence has been grammatically corrected; however, the revision has led to a change in the sentence's meaning. 
    \item \textbf{POOR} The grammatical errors within the sentence remain unaddressed.
\end{itemize}

If both evaluators assign the same quality marker to a sentence, that marker is designated as the final evaluation. Conversely, if their assessments diverge, the third evaluator joins the discussion to reach a consensus, resulting in a definitive quality marker. During the evaluation process, We pay the evaluators at a rate of \$10 for every 100 sentences evaluated, during the discussion process, we pay \$0.5 for each sentence.

\begin{figure}[!htbp]
  \centering
  \includegraphics[width=0.6\linewidth]{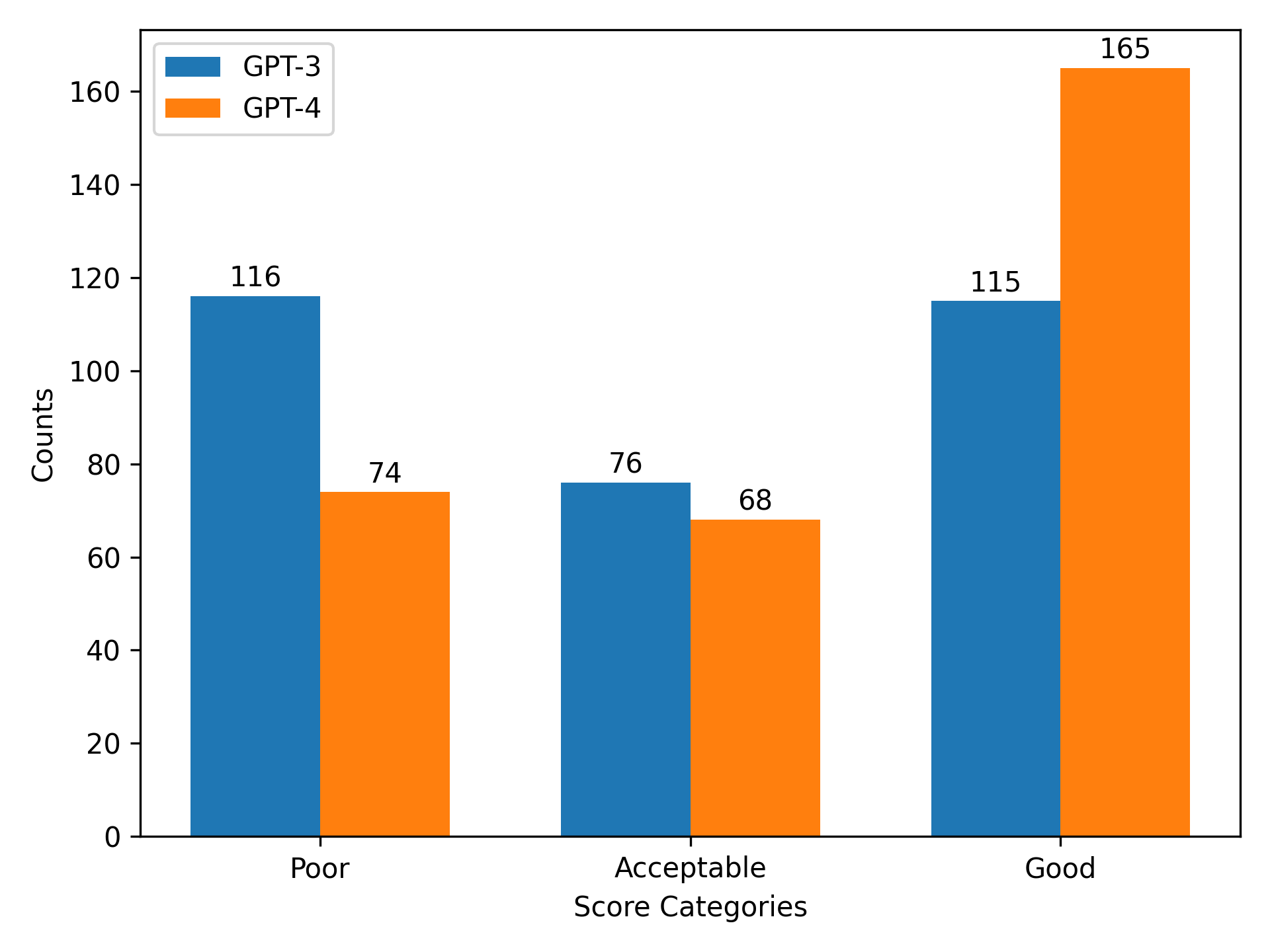}
  \caption{Human Evaluation of the correction of sentences.}
  \label{fig:correct_res}
\end{figure}

\paragraph{Overall Evaluation} Our evaluation of the overall quality of sentence corrections by GPT-3.5 and GPT-4, as presented in Figure \ref{fig:correct_res}, reveals that GPT-4 outperforms GPT-3.5 in terms of correction effectiveness. Notably, GPT-4 produces a higher quantity of accurately corrected sentences, while simultaneously reducing the incidence of sentences rated as poor. This improvement underscores GPT-4's advanced capabilities in understanding and rectifying errors, leading to more reliable and grammatically sound corrections.

\begin{figure}[!htbp]
  \centering
  \includegraphics[width=0.8\linewidth]{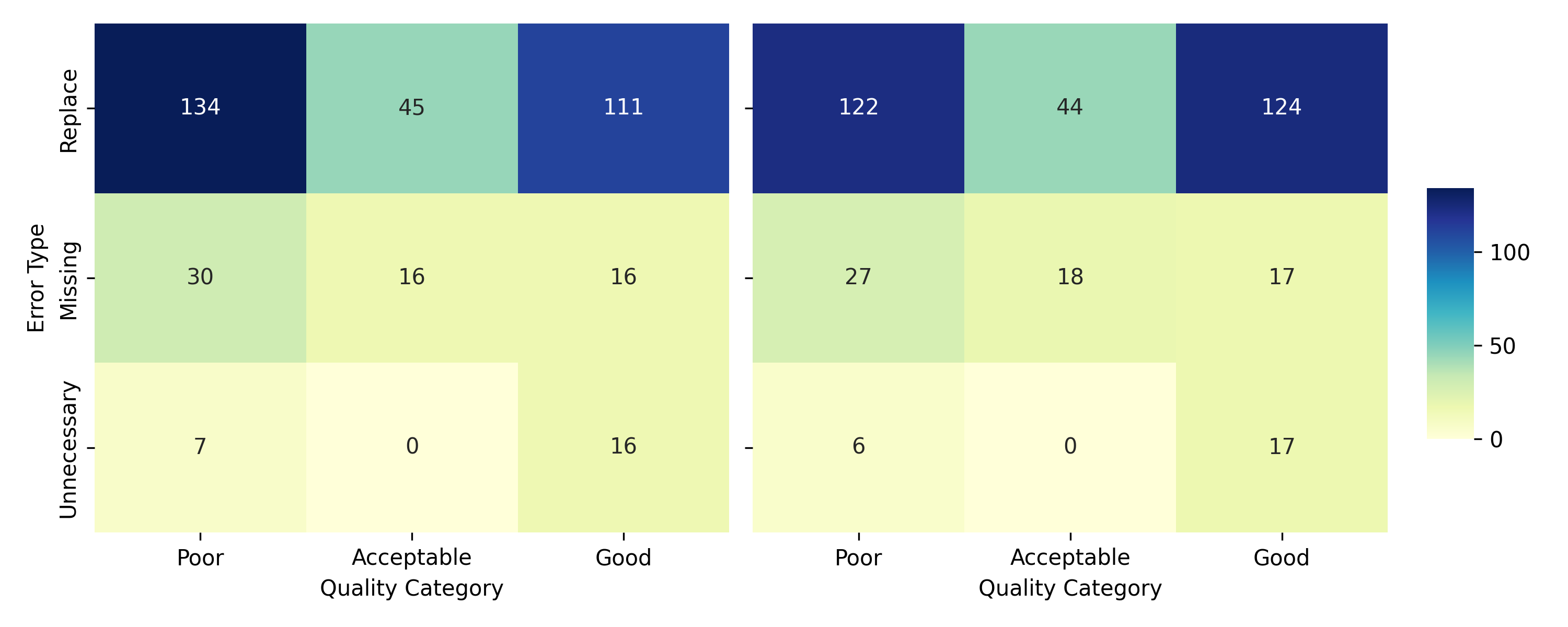}
  \caption{Human Evaluation of Replacement, Unnecessary, and Missing Errors: Comparative Analysis of GPT-3.5 (Left) and GPT4 (Right). The table lists the quality ratings (Good, Acceptable, Poor) for each error type, with results visually differentiated using a color-coded scheme.}
  \label{fig:rmu_res}
\end{figure}

\paragraph{Replace, Missing and Unnecessary Errors} Our analysis further explores corrections across replacement, unnecessary, and missing errors between 0-shot GPT-3.5 and GPT-4, as depicted in Figure \ref{fig:rmu_res}. Both GPT-3.5 and GPT-4 demonstrate proficiency in addressing replacement errors, with over half of such corrections classified as acceptable or good by our evaluators. However, correcting missing errors poses a greater challenge; nearly half of these errors are inadequately addressed, as per our human evaluators' assessments. This particular challenge in correcting missing errors may be attributed to the intrinsic behavior of ChatGPT models, which prioritize contextual coherence and fluency over strict adherence to grammatical rules. LLMs, by design, generate tokens by predicting the most likely next word based on the preceding context. This approach, while effective in producing contextually relevant and coherent text, can sometimes lead to the insertion of tokens that do not strictly conform to grammatical norms.  

Additionally, GPT-4 demonstrates markedly superior performance compared to GPT-3.5 in rectifying all three error types (replacement, unnecessary, and missing). The most pronounced disparity in effectiveness is observed in the correction of replacement errors. Notably, GPT-4 not only produces a higher proportion of corrections classified as "Good" but also significantly reduces the instances of corrections deemed "Poor." This improvement suggests that GPT-4 has a more nuanced understanding of language context and structure, enabling it to more accurately identify and correct errors without compromising the original meaning or grammatical integrity of the sentence. The advanced algorithms and larger dataset on which GPT-4 is trained likely contribute to its enhanced capability to discern and amend errors more effectively, thus leading to corrections that align more closely with the grammatical rules and intended meanings.

\begin{figure}[!htbp] 
  \centering
  \includegraphics[width=0.8\linewidth]{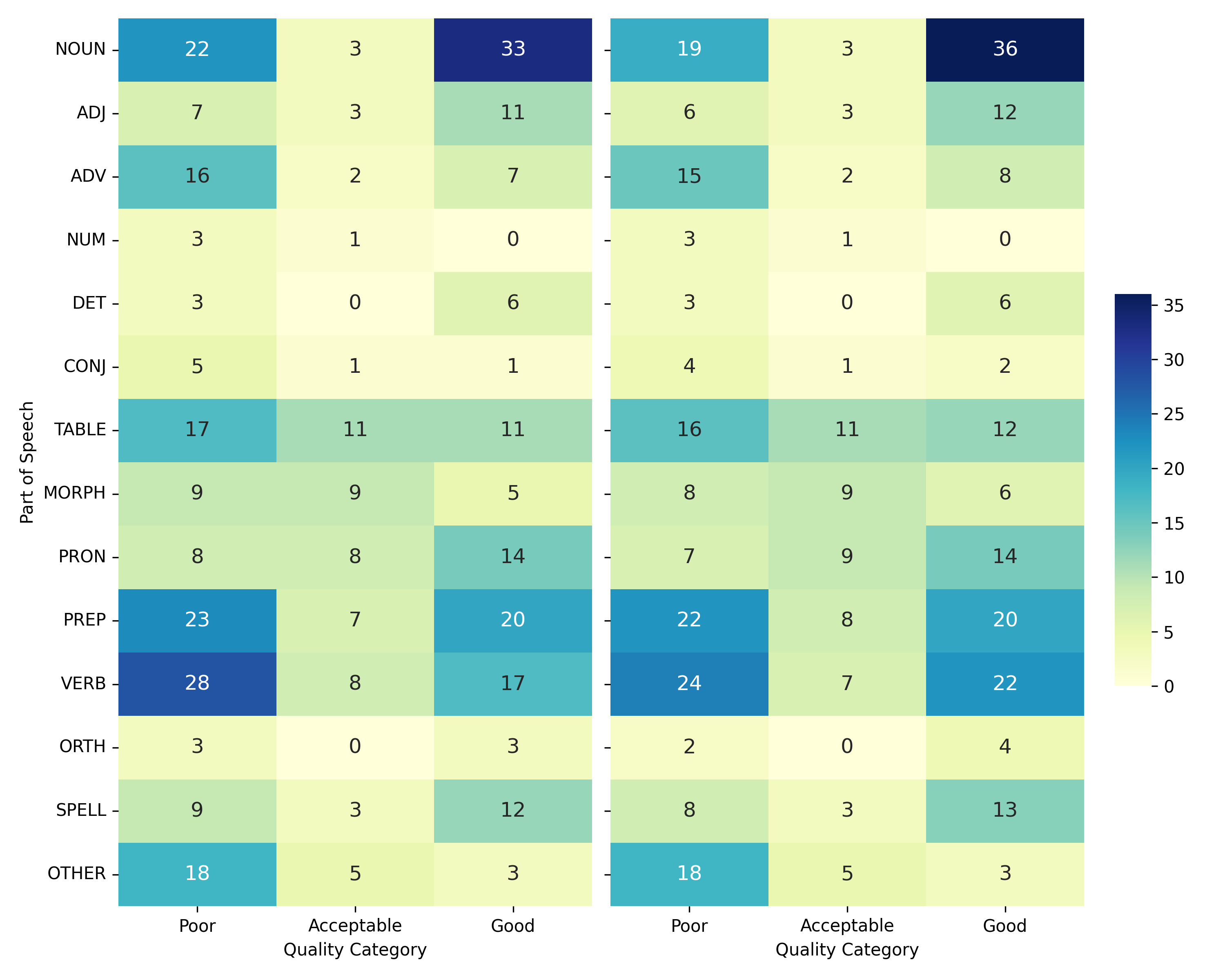}
  \caption{Human evaluation on part of speech errors: a comparative analysis of GPT-3.5 (Left) and GPT-4 (Right). The table lists the quality ratings (Good, Acceptable, Poor) for each error type, with results visually differentiated using a color-coded scheme.}
  \label{fig:pos_res}
\end{figure}

\paragraph{Part of Speech Errors} We now examine the correction of part of speeches listed in Figure \ref{fig:pos_stat}. The correction result is shown in Figure \ref{fig:pos_res}, we observe that the quality of GPT-4 correction results is better than GPT-3.5. 

Regarding part of speeches, there are several observations.  

\begin{itemize}
    \item \textbf{Nouns and Verbs} Noun and verb errors constitute the most prevalent types of part-of-speech errors, overshadowing other error categories. Interestingly, both GPT-3.5 and GPT-4 demonstrate a superior ability to correct noun errors as opposed to verb errors. This discrepancy can be attributed to the relative simplicity of noun variations in Esperanto compared to those of verbs. Noun corrections typically involve adjustments in case, number, and the application of prefixes or suffixes—a process simplified by the more straightforward and consistent grammatical structure of nouns. In contrast, verb corrections often entail the identification and rectification of form and tense errors, presenting a greater challenge for ChatGPT due to the complex nature of verb conjugation and usage in Esperanto. Apart from this, in Figure \ref{fig:pos_res} we reveal that case and number errors can be corrected better than other errors.

    \item \textbf{Adjectives and Adverbs} Additionally, while adjectives and adverbs are crucial elements of Esperanto, our system demonstrated a higher success rate in correcting adjective errors compared to adverbial errors. This difference can be attributed to the greater flexibility inherent to adverbs in Esperanto. Specifically, a noun indicating time or location can be transformed into an adverb simply by removing the noun's terminal "o" and appending the adverbial ending "e". Such morphological flexibility presents unique challenges in error detection and correction for adverbs. Adjective errors, by comparison, are more straightforward for the system to identify and rectify.
    
    \item \textbf{Prepositions and Pronouns} We have observed that LLM is more adept at correcting errors related to pronouns than those associated with prepositions. This discrepancy likely stems from the relatively simple vocabulary and grammatical rules governing pronouns in Esperanto. With only ten personal pronouns (mi, ni, vi, (ci), li, ŝi, ĝi, ili, oni, si) and straightforward grammatical constructions, the rules for pronouns are easily discernible for GPT. Conversely, prepositions pose a greater challenge. The complexity arises not only from their varied usage but also from the subtleties in meaning that different prepositions convey, making them more difficult for the model to accurately correct.
\end{itemize}

\begin{figure}[!htbp] 
  \centering
  \includegraphics[width=0.8\linewidth]{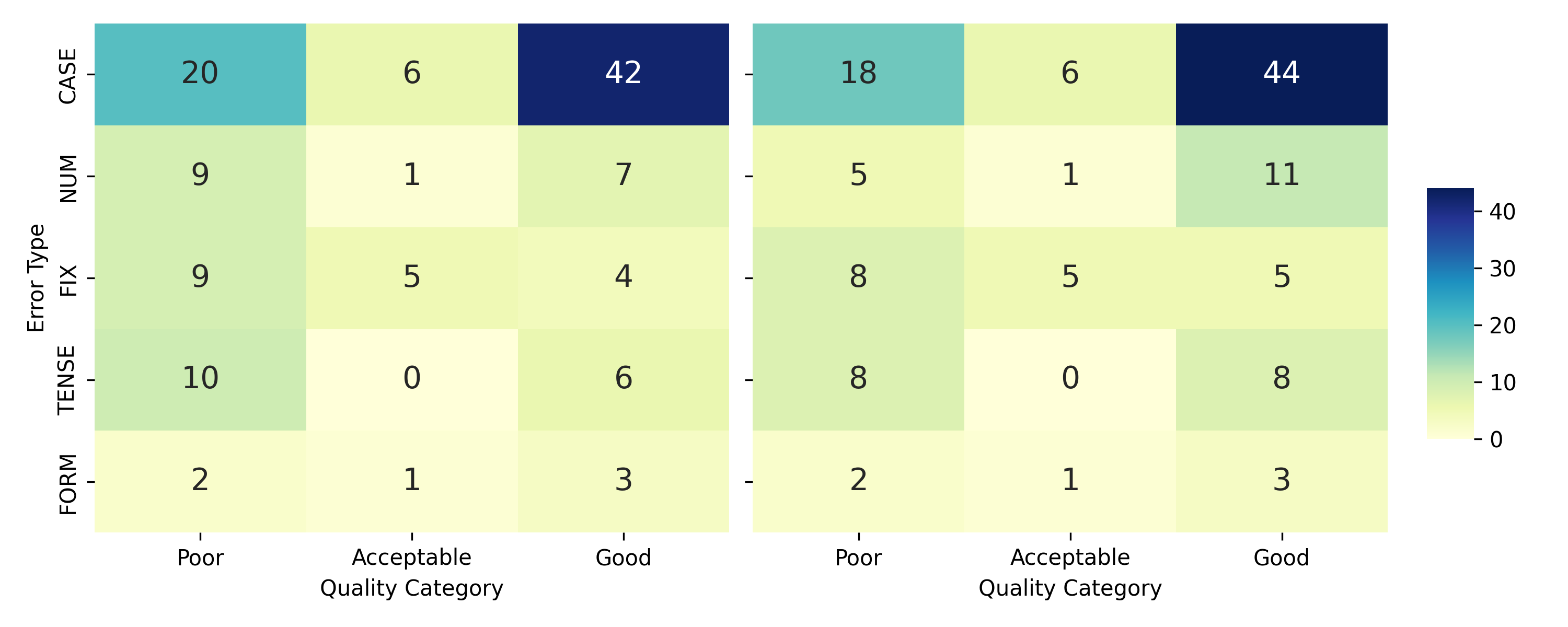}
  \caption{Human evaluation on five grammatical error types: comparative analysis of GPT-3.5 (Left) and GPT-4 (Right). The table lists the quality ratings (Good, Acceptable, Poor) for each error type, with results visually differentiated using a color-coded scheme.}
  \label{fig:error_type_res}
\end{figure}

\paragraph{Grammar Errors} In this section, we delve into the models' proficiency in correcting various types of grammatical errors under 0-shot scenario, specifically CASE, NUM, FIX, TENSE, and FORM. The outcomes are illustrated in Figure \ref{fig:error_type_res}. Across the board, GPT-4 outperforms GPT-3.5, showcasing a notably higher efficiency in rectifying errors. Notably, both models exhibit a stronger capability in correcting CASE errors compared to other types. This superiority is likely attributable to Esperanto's stringent case rules, which are consistently applied and, therefore, more straightforward for the models to learn and apply accurately.

\section{Case Study}
In this part, we analyze the cases of grammar error correction using LLMs under 0-shot scenario. We list 8 cases in Table \ref{tab:Case_Studies}. Overall, we observe that GPT-4 has a better correction ability than GPT-3.5. Below are detailed observations. 
\begin{itemize}
    \item \textbf{Tense of Verbs} GPT-4 demonstrates a heightened sensitivity to verb tense within complex sentences, particularly in scenarios invoking the conditional mood. This is evident in Examples 1 and 3, where the tense of a verb critically hinges on the likelihood of the action's occurrence: actions deemed unlikely necessitate the conditional mood, whereas the verb tense should align with temporal context when the action has a higher likelihood of occurrence. GPT-4 adeptly discerns this distinction. For instance, "Se ne \textcolor{red}{pluvas} morgaŭ (If it doesn't rain tomorrow)" is a future-oriented statement, thus accurately revised by GPT-4 to "Se ne \textcolor{blue}{pluvos} morgaŭ", reflecting the appropriate future tense.
    \item \textbf{Noun Phrases Used as Adverbs} In Example 2, the phrase "\textcolor{red}{ĉiujn jaroj}" (every years) is grammatically incorrect and should be amended to "\textcolor{blue}{ĉiun jaron}" (every year) to accurately reflect Esperanto grammar rules. The correction to an 'n' ending is required because the phrase functions as a temporal adverbial phrase, denoting a duration of time that necessitates the accusative case. This change not only aligns with the linguistic standards of Esperanto but also clarifies the intended meaning of the phrase as referring to a recurring annual event.
    \item \textbf{Suffixes} In Example 4, the correction from "\textcolor{red}{trinkenda}" (must to drink) to "\textcolor{blue}{trinkinda}" (worth to drink) is necessary to accurately convey the intended meaning. Similarly, Example 7 requires an adjustment from "\textcolor{red}{malkonstruigitaj}" to "\textcolor{blue}{malkonstritaj}" to indicate that the action of tearing down the wall was performed by herself, not by others. The addition of the "-ig" suffix mistakenly implies an active, causative role by the subject, which is not the case. These examples underscore the agglutinative nature of Esperanto, where prefixes and suffixes play a pivotal role in defining the sentence's meaning. The correct use of these grammatical elements is essential for conveying precise nuances and ensuring the sentence accurately reflects the speaker's intent.

    \item \textbf{Spelling Errors and Similar Words} GPT-4 and GPT-3.5 do not successfully correct the spelling mistake "\textcolor{red}{regjono}," which should be spelled "\textcolor{blue}{regiono}" in Esperanto. This oversight suggests that LLMs may grasp the intended meaning of "regjono" despite the non-standard suffix “-jono.” Such instances highlight the models' capacity to understand contextually correct meanings, even when faced with orthographic inaccuracies. However, this quality is not desired when used for Grammar Error Correction
    \item \textbf{Nuances in Grammar} In Example 8, the preposition "\textcolor{red}{malantaŭ}" is incorrectly employed to denote the temporal or sequential relationship between two events, a function it cannot fulfill in Esperanto. The appropriate preposition for expressing such relationships is "\textcolor{blue}{post}." Unfortunately, both GPT-4 and GPT-3.5, do not identify and correct this error. This oversight underscores a specific challenge in preposition usage within language models, highlighting a gap in their ability to discern and apply the correct prepositional terms based on contextual cues related to time and sequence.
\end{itemize}

\input{tables/case}

\section{Conclusion}
In conclusion, this study addresses the gap in Grammar Error Correction (GEC) for low-resource languages like Esperanto, leveraging tailored datasets (Eo-GP and Eo-GEC) and advanced models (GPT-3.5 and GPT-4). The superior performance of GPT-4, demonstrated through both automated and human evaluations, highlights its potential to correct Esperanto's grammatical errors. However, the limitation of data volume underscores the need for expanded datasets to deepen future research. This research aims to serve as an impetus for further research into the influences of LLMs on Esperanto as well as other non-English linguistic systems.

\appendix
\section{Esperanto Alphabet}\label{sec:eo-alphabet}
A complete Esperanto alphabet is shown in Table \ref{tab:esperanto_alphabet}. Notice that when compared to English, Esperanto does not have q,w,x, and y but includes six letters that have diacritical marks: ĉ, ĝ, ĥ, \^{j}, ŝ, and ŭ.
\input{tables/eo_alphabet}

\section{Part of Speech of Esperanto Words} 
\input{tables/part_of_speech}
The part of speech for Esperanto words are listed in Table \ref{tab:pos}. We notice that different grammar textbooks have slight differences in the definitions of parts of speech. Based on this, we define the following parts of speech by combining the annotation characteristics of the grammatical error correction task with the foundation of \textit{Plena Manlibro de Esperanta Gramatiko} (PMEG, Complete Manual of Esperanto Grammar) \footnote{\url{https://bertilow.com/pmeg/index.html}}.

\section{A Complete list of valid error code combinations}\label{apdx:list_of_error_type}
A comprehensive error code combination of Esperanto grammatical error correction for POS, morphology and other errors is presented in Table \ref{tab:error_pos},\ref{tab:error_morph},\ref{tab:error_other}. N/A signifies that this type of error does not exist.
\input{tables/all_error_types_POS}
\input{tables/all_error_types_morph}
\input{tables/all_error_types_other}

\newpage
\bibliographystyle{unsrt}  
\bibliography{references}

\end{document}

%% file: tables/zipf_fitting.tex
\begin{table}
  \caption{Fitting of Zipf's law on all Esperanto words' frequency and non-stop words' frequency, where $k$ and $a$ are parameters in equation \ref{eqn:zipf-mandelbrot-log}, while $E_s$ means standard error}
  \label{tab:zipf_fitting}
  \begin{center}

  \begin{tabular}{ccc}
    \toprule
    Parameters&All words&Non-stop words\\
    \midrule
    $\log k$ & 11.71 & 9.67 \\
    $a$ & 0.88 & 0.62\\
    $R^2$ & 0.99 & 0.98\\
    $E_s$ & 7.57$\times 10^{-3}$ & 8.49$\times 10^{-3}$ \\
  \bottomrule
\end{tabular}
\end{center}
\end{table}

%% file: tables/example_of_errors.tex
\begin{table}
    \small
  \caption{A list of main error categories in our framework with examples and explanations.}
  \label{tab:example_of_errors}
  \begin{tabular}{l|ll}
  \toprule
    Code & Meaning & Description/Example\\
    \midrule
    R:VERB:SVA & Subject-Verb Agreemet		&	Mi \textcolor{red}{loĝi} $\rightarrow$ \textcolor{blue}{loĝas} en Romo. \\
    R:VERB:TENSE & Verb tense & Includes inflectional and periphrastic tense, modal verbs and passivization. \\
    R:VERB:FORM & Verb form & Inifinitives, gerunds and participles. devas \textcolor{red}{estas} $\rightarrow$ devas \textcolor{blue}{esti} \\ 
    R:PREP & Preposition & \textcolor{red}{al} $\rightarrow$ \textcolor{blue}{en} \\
    R:SPELL & Spelling & \textcolor{red}{regjono} $\rightarrow$ \textcolor{blue}{regiono} \\
    R:TABLE & Table Words & \textcolor{red}{kiu} $\rightarrow$ \textcolor{blue}{kie} \\
    R:MORPH & Morphology & Token have similar lemma but nothing else in common. ĉiu \textcolor{red}{tago} $\rightarrow$ ĉiu \textcolor{blue}{tage} \\
    R:PRON & Pronouns & \textcolor{red}{sia} frato $\rightarrow$ \textcolor{blue}{lia} frato \\
    R:NOUN:FIX & Prefix or suffix of nouns & \textcolor{red}{filoj} $\rightarrow$ \textcolor{blue}{gefiloj} \\
 
  \bottomrule
\end{tabular}
\end{table}

%% file: tables/correction_quality.tex
\begin{table}
\begin{center}

  \caption{The correction quality evaluated by ERRANT and M2Scorer}
  \label{tab:correction_res}
  \begin{tabular}{cccc|ccc}
  \toprule 
           & \multicolumn{3}{c}{ERRANT} & \multicolumn{3}{c}{M2Scorer} \\
    \midrule
Method & $P$       & $R$       & $F_{0.5}$   & $P$        & $R$       & $F_{0.5}$    \\
GPT-3.5 0-shot         & 42.22   & 32.90    & 39.96  & 43.90     & 35.61   & 41.94   \\
GPT-3.5 1-shot         & 47.41   & 51.02   & 48.09  & 49.57    & 40.57   & 47.46   \\
GPT-3.5 3-shot       & 50.47   & \textbf{56.76}   & 51.62  & 52.43    & 38.21   & 48.80    \\
GPT-3.5 5-shot         & \textbf{50.94}   & 56.25   & \textbf{51.92}  & 50.96    & 37.50    & 47.55   \\
\midrule
GPT-4 0-shot          & 38.92   & 45.08   & 40.01  & 53.23    & 48.58   & 52.23  \\
GPT-4 1-shot & 37.50 &	44.41 &	38.70 &	\textbf{54.12} & \textbf{49.53} &	\textbf{53.14} \\
GPT-4 3-shot & 41.27 &	46.79 &	42.27 &	53.78 &	46.93 &	52.26 \\
GPT-4 5-shot & 42.22 &	47.73 &	43.22 &	53.97 &	46.46 &	52.28 \\
  \bottomrule
\end{tabular}
\end{center}
\end{table}

%% file: tables/case.tex
\begin{table}[t]
\small
\footnotesize
\centering
\begin{tabular}{ll}
\toprule


\textbf{Example 1:} & Se ne \textcolor{red}{pluvas} morgaŭ, ni \textcolor{red}{povas} ludi tenison. \\ 
\textbf{Target:} &  Se ne \textcolor{blue}{pluvos} morgaŭ, ni \textcolor{blue}{povos} ludi tenison. \\ & (If it doesn't \textbf{rain} tomorrow, we can \textbf{play} tennis.) \\
\textbf{GPT-3.5:} & Se ne \textcolor{red}{pluvas} morgaŭ, ni \textcolor{red}{povas} ludi tenison. \\
\textbf{GPT-4} & Se ne \textcolor{blue}{pluvos} morgaŭ, ni \textcolor{red}{povas} ludi tenison. \\ \midrule \midrule

\textbf{Example 2:} & \textcolor{red}{Ĉiunj} \textcolor{red}{jaroj} ni vizitas nian fratinon en Aŭstralio. \\ 
\textbf{Target:} &  \textcolor{blue}{Ĉiun} \textcolor{blue}{jaron} ni vizitas nian fratinon en Aŭstralio.\\ & (\textbf{Every} \textbf{year}  we visit our sister in Australia.) \\
\textbf{GPT-3.5:} & \textcolor{red}{Ĉiujn} \textcolor{red}{jarojn} ni vizitas nian fratinon en Aŭstralio. \\
\textbf{GPT-4} & \textcolor{blue}{Ĉiun} \textcolor{blue}{jaron} ni vizitas nian fratinon en Aŭstralio. \\ \midrule \midrule

\textbf{Example 3:} & Se mi scius, ke ŝi ne \textcolor{red}{venus}, ankaŭ mi ne venus.
 \\ 
\textbf{Target:} &  Se mi scius, ke ŝi ne \textcolor{blue}{venos}, ankaŭ mi ne venus.\\ & (If I knew she wouldn't \textbf{come}, I wouldn't come either.) \\
\textbf{GPT-3.5:} & Se mi \textcolor{orange}{sciis}, ke ŝi ne \textcolor{red}{venus}, ankaŭ mi ne venus.
\\
\textbf{GPT-4} &Se mi scius, ke ŝi ne \textcolor{blue}{venos}, ankaŭ mi ne venus. \\ \midrule \midrule

\textbf{Example 4:} & La teo ĉi tie estas vere \textcolor{red}{trinkenda}, laŭ mi.
 \\ 
\textbf{Target:} &  La teo ĉi tie estas vere \textcolor{blue}{trinkinda}, laŭ mi.
\\ & (The tea here is really \textbf{worth to drink}, in my opinion.) \\
\textbf{GPT-3.5:} & La teo ĉi tie estas vere \textcolor{orange}{trinkebla}, laŭ mi.
\\
\textbf{GPT-4} &La teo ĉi tie estas vere \textcolor{blue}{trinkinda}, laŭ mi.
 \\ \midrule \midrule

 \textbf{Example 5:} & La bela \textcolor{red}{regjono} de Esperanto estas la tuta mondo, ĉar ĝi unuigas homojn el ĉiuj kulturoj kaj landoj.
 \\ 
\textbf{Target:} & La bela \textcolor{blue}{regiono} de Esperanto estas la tuta mondo, ĉar ĝi unuigas homojn el ĉiuj kulturoj kaj landoj.
\\ & (The beautiful \textbf{region} of Esperanto is the whole world, because it unites people from all cultures and countries.) \\
\textbf{GPT-3.5:} &La bela \textcolor{red}{regjono} de Esperanto estas la tuta mondo, ĉar ĝi unuigas homojn el ĉiuj kulturoj kaj landoj.
\\
\textbf{GPT-4} &La bela \textcolor{orange}{reĝino} de Esperanto estas la tuta mondo, ĉar ĝi unuigas homojn el ĉiuj kulturoj kaj landoj.
 \\ \midrule \midrule

\textbf{Example 6:} & Mi \textcolor{red}{estrigis} de mia laborposteno, antaux ol mia nova estro povos \textcolor{red}{eksigis} min!
        \\ 
\textbf{Target:} & Mi \textcolor{blue}{eksiĝis} de mia laborposteno, antaux ol mia nova estro povos \textcolor{blue}{eksigi} min!
\\ & (I \textbf{quit} my job before my new boss can \textbf{fire} me!) \\
\textbf{GPT-3.5:} &Mi \textcolor{orange}{retiriĝis} de mia laborposteno, antaŭ ol mia nova estraro \textcolor{orange}{povas} \textcolor{blue}{eksigi} min!
\\
\textbf{GPT-4} &Mi \textcolor{blue}{eksiĝis} de mia laborposteno, antaŭ ol mia nova estro povos \textcolor{blue}{eksigi} min!
 \\ \midrule \midrule

 \textbf{Example 7:} & La muroj estis \textcolor{red}{malkonstruigitaj} de ŝi mem.
        \\ 
\textbf{Target:} & La muroj estis \textcolor{blue}{malkonstruitaj} de ŝi mem.
\\ & (The walls were \textbf{torn down} by herself.) \\
\textbf{GPT-3.5:} &La muroj estis \textcolor{blue}{malkonstruitaj} de \textcolor{orange}{si} mem.
\\
\textbf{GPT-4} &La muroj estis \textcolor{blue}{malkonstruitaj} de ŝi mem.
 \\ \midrule \midrule

\textbf{Example 8:} & Li erare metis la fina\^{j}on N \textcolor{red}{malantaŭ} tiu vorto.
\\ 
\textbf{Target:} & Li erare metis la fina\^{j}on N \textcolor{blue}{post} tiu vorto.
\\ & (He mistakenly put the ending N \textbf{after} that word.) \\
\textbf{GPT-3.5:} &Li erare metis la fina\^{j}on \textcolor{orange}{n} \textcolor{red}{malantaŭ} tiun \textcolor{orange}{vorton}.
\\
\textbf{GPT-4} &Li erare metis la fina\^{j}on \textcolor{orange}{-n} \textcolor{red}{malantaŭ} tiu vorto.
\\ \bottomrule

\end{tabular}
\caption{Case Study for Esperanto Grammar Correction on GPT-3.5 and GPT-4, the English translation of the original sentence is listed and the corresponding English translation of the wrong token is marked with \textbf{bold}. We mark the \textcolor{red}{wrong tokens}/\textcolor{orange}{wrong edits}/\textcolor{blue}{correct edits} in \textcolor{red}{red}/\textcolor{orange}{orange}/\textcolor{blue}{blue}. }
\label{tab:Case_Studies}
\end{table}

%% file: tables/eo_alphabet.tex
\begin{table}[!htb]
\centering
\caption{Esperanto Alphabet}
\label{tab:esperanto_alphabet}
\begin{tabular}{cc|cc|cc}
\toprule
Uppercase & Lowercase & Uppercase & Lowercase & Uppercase & Lowercase \\
\midrule
A & a & Ĥ & ĥ & P & p \\
B & b & I & i & R & r \\
C & c & J & j & S & s \\
Ĉ & ĉ & Ĵ & \^{j}  & T & t \\
D & d & J & j & U & u \\
E & e & K & k & Ŭ & ŭ  \\
F & f & L & l & V & v  \\
G & g & M & m & Z & z \\
Ĝ & ĝ & N & n &  &  \\
H & h & O & o &  &  \\
\bottomrule
\end{tabular}
\end{table}

%% file: tables/part_of_speech.tex
\begin{table}[!htb]
  \caption{Part of Speech in Esperanto, the corresponding word in Esperanto is marked in \textcolor{red}{red}, and the corresponding translation in English is marked in \textbf{bold}. It has been observed that the translation between English and Esperanto is not strictly one-to-one. The English translation might omit certain words due to grammar forms. When an English verb functions as a direct transitive verb (e.g., 'give'), the translation omits the English equivalent of the Esperanto preposition 'al'.}
  \label{tab:pos}
  \begin{center}

  \begin{tabular}{ccc}
    \toprule
    Part of speech & Example & Translation\\
    \midrule
    Noun &  Mi vidas \textcolor{red}{leonon} & I see a \textbf{lion}. \\
    Adjective & Tiuj ĉi verkistoj estas \textcolor{red}{famaj}. & These writers are \textbf{famous}. \\
    Adverb &  Ili manĝas \textcolor{red}{rapide}. & They eat \textbf{quickly}. \\
    Number & \textcolor{red}{Tri} grandaj domoj & \textbf{Three} large houses \\
    Determiner & \textcolor{red}{La} nokton ŝi pasigis ĉe sia laboro. & She worked \textbf{the} whole night \\
    Preposition & La reĝidino donis \textcolor{red}{al} li glavon. & The princess gave him a sword. \\
    Pronoun & \textcolor{red}{Ili} ankaŭ estas en la ĝardeno. &  \textbf{They} are also in the garden. \\
    Conjunction &  Petro manĝas per forko \textcolor{red}{kaj} tranĉilo. & Peter eats with a fork \textbf{and} knife. \\
  \bottomrule
\end{tabular}
      
\end{center}
\end{table}

%% file: tables/all_error_types_POS.tex
\begin{table}[!htb]
  \caption{Error Combination of different part of speeches}
  \label{tab:error_pos}
  \begin{center}
  \begin{tabular}{l|lll}
  \toprule
    Type & Missing & Unnecessary & Replacement \\
    \midrule
    Adjective &		M:ADJ&	U:ADJ&	R:ADJ \\
    Adverb &		M:ADV&	U:ADV&	R:ADV \\
    Conjunction &		M:CONJ&	U:CONJ&	R:CONJ \\
    Determiner &		M:DET&	U:DET&	R:DET \\
    Noun &		M:NOUN&	U:NOUN&	R:NOUN\\
    Preposition &		M:PREP&	U:PREP&	R:PREP\\
    Pronoun &		M:PRON&	U:PRON&	R:PRON\\
    Punctuation &		M:PUNCT &	U:PUNCT &	R:PUNCT \\
    Verb &		M:VERB &	U:VERB &	R:VERB  \\
    Table Words & M:TABLE & U:TABLE & R:TABLE \\
  \bottomrule
\end{tabular}
\end{center}
\end{table}

%% file: tables/all_error_types_morph.tex
\begin{table}[!htb]
  \caption{Error Combination of different morphology levels}
  \label{tab:error_morph}
  \begin{center}
    \begin{tabular}{ll|lll}
    \toprule
        Type & Error & Mission & Unnessary & Replace\\ \midrule
        \multirow{4}{*}{Adjective} & Case & N/A & N/A & R:ADJ:CASE \\
        & Number & N/A & N/A & R:ADJ:NUM \\ 
        & Prefix/Suffix & N/A & N/A & R:ADJ:FIX \\ 
        \midrule
        \multirow{4}{*}{Noun} & Case & N/A & N/A & R:NOUN:CASE \\ 
        & Number & N/A & N/A & R:NOUN:NUM \\ 
        & Prefix/Suffix & N/A & N/A & R:NOUN:FIX \\ 
        \midrule
        \multirow{3}{*}{Verb} & FORM & N/A & N/A & R:VERB:FORM \\ 
        & SVA & N/A & N/A & R:VERB:SVA \\ 
        & Tense & N/A & N/A & R:VERB:TENSE \\ 
        \midrule
        Adverb & Case & N/A & N/A & R:ADV:CASE \\ 
        \midrule
        \multirow{2}{*}{Pronoun} & Case & N/A & N/A & R:PRON:CASE \\ 
        & Number & N/A & N/A & R:PRON:NUM \\
        \midrule
        \multirow{2}{*}{Table Words} & Case & N/A & N/A & R:TABLE:CASE \\
        & Number & N/A & N/A & R:TABLE:NUM \\
        \bottomrule
    \end{tabular}
    \end{center}
\end{table}

%% file: tables/all_error_types_other.tex
\begin{table}[!htb]
  \caption{Error Combination of Non-Part-of-Speech Error Types}
  \label{tab:error_other}
  \begin{center}
  \begin{tabular}{l|lll}
  \toprule
    Type & Missing & Unnecessary & Replacement \\
    \midrule
    Morphology&	N/A&	N/A&	R:MORPH\\
    Orthography&	N/A&	N/A&	R:ORTH\\
    Other&	M:OTHER&	U:OTHER&	R:OTHER\\
    Spelling&	N/A&	N/A&	R:SPELL\\
    Word Order&	N/A&	N/A&	R:WO\\
  \bottomrule
\end{tabular}
\end{center}
\end{table}